\documentclass[10pt,twocolumn,letterpaper]{article}

\usepackage{cvpr}
\usepackage{times}
\usepackage{epsfig}
\usepackage{graphicx}
\usepackage{amsmath}
\usepackage{amssymb}
\usepackage{comment}
\usepackage{chngcntr}
\usepackage[title]{appendix}

\usepackage[pagebackref=true,breaklinks=true,letterpaper=true,colorlinks,bookmarks=false]{hyperref}
\cvprfinalcopy 

\begin{document}

\title{Convolutional Neural Networks Deceived by Visual Illusions}

\author{A. Gomez-Villa\thanks{Department of Information and Communication Technologies (DTIC), Universitat Pompeu Fabra. C. Roc Boronat 138, 08018, Barcelona, Spain}, A. Mart\'in\footnotemark[1],
J. Vazquez-Corral\thanks{CMP, University of East Anglia,NR4 7TJ, Norwich, UK},
 M. Bertalm\'io\footnotemark[1]\\
{\tt\small \{alexander.gomez, adrian.martin, marcelo.bertalmio\}@upf.edu, j.vazquez@uea.ac.uk}
}

\maketitle

\begin{abstract}

Visual illusions teach us that what we see is not always what it is represented in the physical world.
Its special nature make them a fascinating tool to test and validate any new vision model proposed. In general, current vision models are based on the concatenation of linear convolutions and non-linear operations. In this paper we get inspiration from the similarity of this structure with the operations present in Convolutional Neural Networks (CNNs). This motivated us to study if CNNs trained for low-level visual tasks are deceived by visual illusions. In particular, we show that CNNs trained for image denoising, image deblurring, and computational color constancy are able to replicate the human response to visual illusions, and that the extent of this replication varies with respect to variation in architecture  and spatial pattern size. We believe that this CNNs behaviour appears as a by-product of the training for the low level vision tasks of denoising, color constancy or deblurring.  Our work opens a new bridge between human perception and CNNs: in order to obtain CNNs that better replicate human behaviour, we may need to start aiming for them to better replicate visual illusions.

\end{abstract}

\section{Introduction}\label{sec:introduction}

Visual illusions are fascinating examples of the complexity of the human visual system, and of the intrinsic difference between perception and reality: while we constantly assume that what we see is a faithful representation of the world around us, visual illusions make clear that what we see is just an internal construct of eyes and brain, because our internal representation and the world itself often do not match.

\begin{figure}[ht]
\begin{center}
\includegraphics[width=0.9\linewidth]{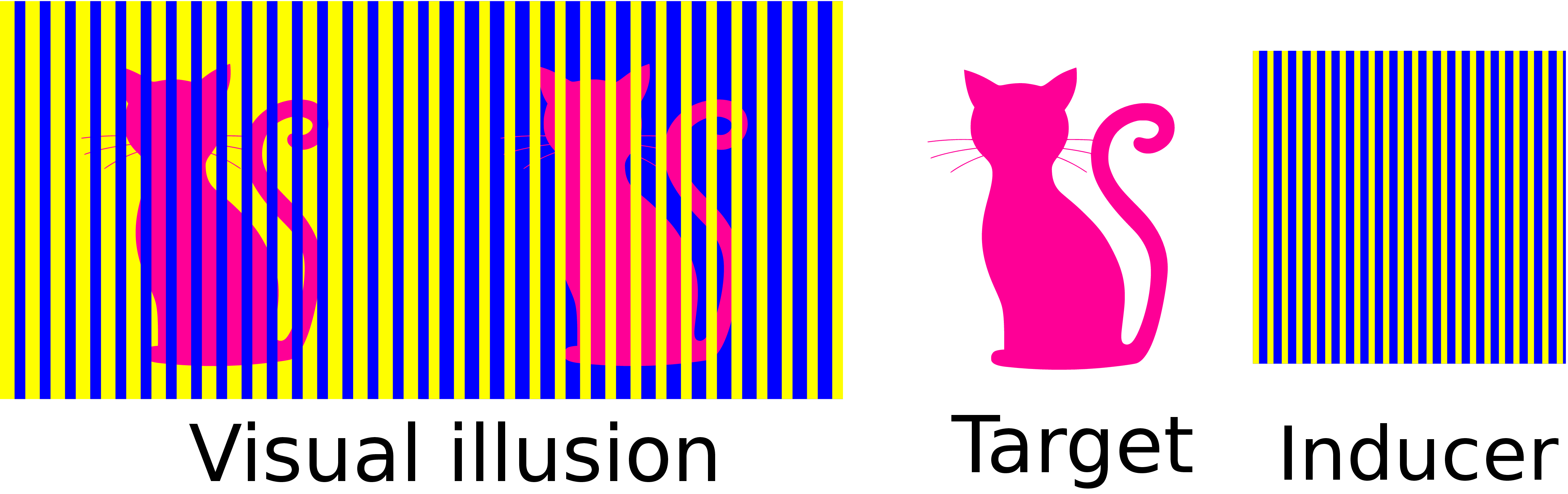}
\end{center}
\caption{Anatomy of a simple color visual illusion. While the target (cat) is always the same, with the same RGB triplet in the three cases, we perceive it as ``pink'' when it is isolated, but magenta with one inducing surround and orange with the other.}
\label{fig:anatomy}
\end{figure}

For instance, Fig.~\ref{fig:anatomy} shows a simple color illusion, where three identical cats are seen as having quite different colors depending on their surround. 
Visual illusions are so striking because, even after we go and check that the three cats have indeed the same triplet RGB value and therefore send the same light to us, we still see them as having different colors.

There are many types of illusions apart from color-based, involving other percepts such as brightness, motion, geometry or grouping, to name a few \cite{shapiro2016oxford}.
For the visual science community the study of visual illusions is key \cite{kingdom2011lightness,kimbertalmioJEI2017}, as the mismatches between reality and perception provide insights that can be very useful to develop new vision models of perception or of neural activity \cite{bookHubel}, and also to validate existing ones. This remains a very challenging open problem, as attested by the variety of vision science models (e.g. perceptual models based on edge-integration, Gestalt-anchoring, spatial-filtering, intrinsic images or purely empirical ones) and the fact that none of them can replicate a wide range of visual illusions; even models that can successfully predict an illusion may fail when a slight modification (like adding noise) is introduced \cite{betz2015noise}.

A very popular approach in vision science is to model neural activity and also perception as a cascade of modules, each consisting of a linear convolution operation followed by a nonlinearity, see \cite{plosone} and references therein\footnote{We want to stress that linear+nonlinear cascades are very common but definitely not the only approach to modeling in vision science, given their well-known limitations.}.
These are of course the building blocks of convolutional neural networks (CNNs), but while the filters in visual models are designed so that the model best replicates neural or perceptual data, filters in CNNs are learned in a supervised manner in order to perform a specific imaging task, such as classification, recognition or denoising, to name just a few.

The authors find rather stunning that, given the importance of visual illusions for the vision science community, the neural inspiration of CNNs, and that so often the aim of CNNs is replicating human behaviour, there is virtually no work done on linking visual illusions and CNNs. To the best of our knowledge there are only two, very recent, publications in this regard. The first one comes from the vision science's field \cite{watanabe2018illusory}, where a CNN trained to predict videos was able to reproduce motion illusions.  In the second one, from computer  vision's perspective \cite{williams2018optical}, the authors classify and attempt to generate new visual illusions using generative adversarial networks.

In this paper we report what we consider to be a quite remarkable and surprising finding, namely that CNNs trained on natural image databases for basic low-level vision tasks reproduce the human response to visual illusion images, i.e. the CNNs are deceived by the visual illusions in the same way that we are deceived by them.
Our other main contribution is to study how the ability of these CNNs to replicate visual illusions is affected by common architecture variations and spatial pattern size.

These results have, we believe, important consequences both for visual science and computer vision. For the visual science community, they support the idea that in order to perform low level vision tasks, the human visual system performs operations that as by-product create visual illusions. Moreover, these findings could help vision science in developing a taxonomy of which visual illusions are associated to which visual tasks. For the computer science community, the results build a new bridge between CNNs and the visual system. However, as it is shown on our experiments, this relationship and its possible consequences are constrained by the fact that not all optical illusions are replicated by the CNNs here studied. This can shed light on the limitations of CNNs for mimicking the visual system, and therefore offers an opportunity for the design of new architectures that, by better replicating visual illusions, could behave more like humans do.

\section{Methods: selected visual illusions and CNNs}\label{sec:methods}

\begin{figure}[ht]
\centering
\includegraphics[width=0.75\linewidth]{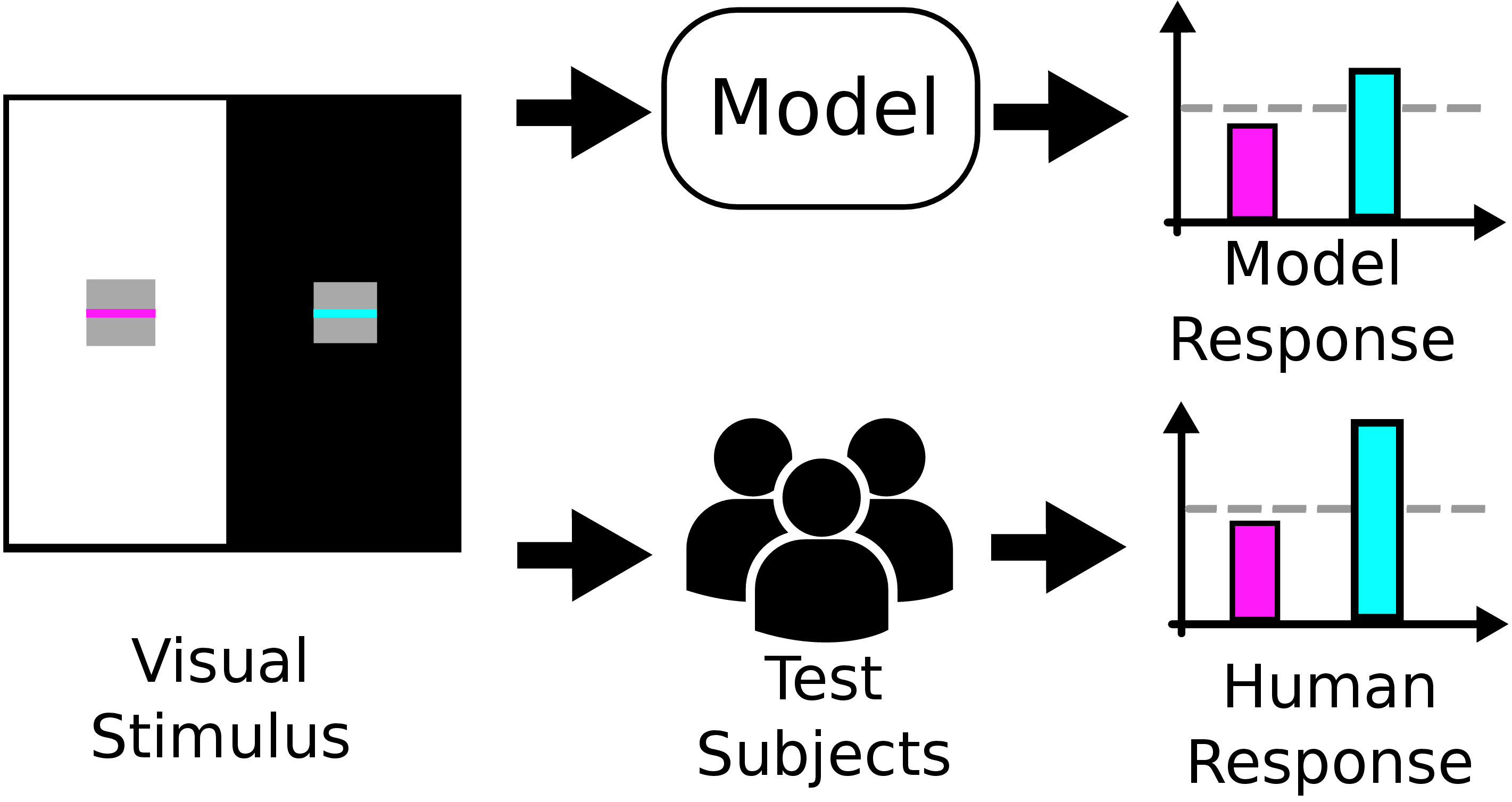}
\caption{Scheme of a classical pipeline to test the replication of a visual illusion by visual models}
\label{fig:visualSciencePipeline}
\end{figure}

Figure \ref{fig:visualSciencePipeline} shows a scheme of the usual procedure to measure the capacity of a model for replicating visual perception in some particular scenario. The observers first assess their perception of some aspect of the stimulus (e.g. the brightness) in a manner that is quantifiable (e.g. by ranking brightness on a scale from 0 to 5), Then, subject responses are averaged, and these averages are compared with the values of the image outputs produced by the model. The better the match, the better the model. This model validation can be performed either just qualitatively or quantitatively too.

As mentioned above, in this paper we will take a few CNNs trained for low-level vision tasks and show that when they are applied to some visual illusion images they produce image outputs that are consistent with our perception.

\begin{figure*}[ht]
\begin{center}
\includegraphics[width=\linewidth]{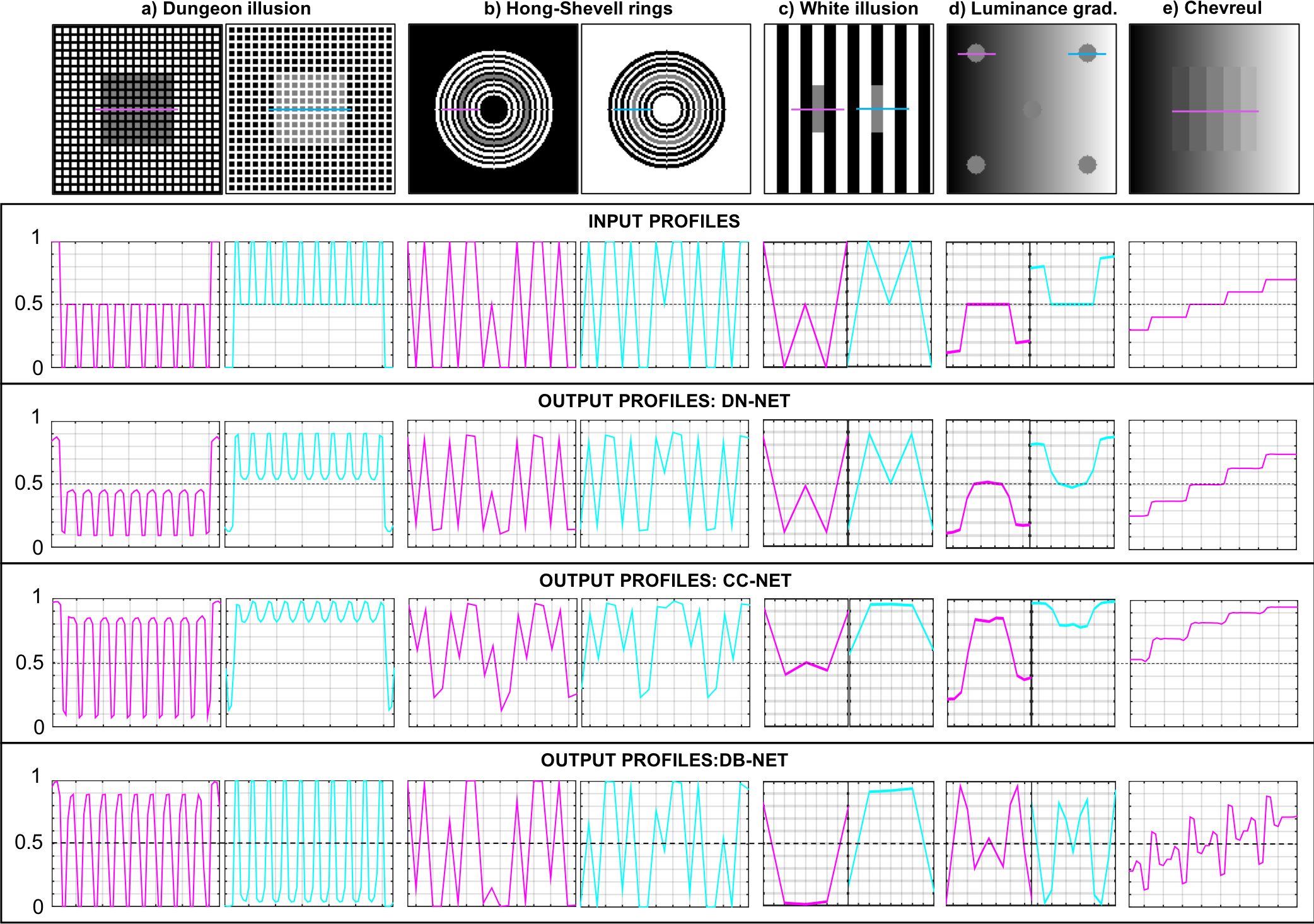}
\end{center}
   \caption{The first row displays the selected grayscale visual illusions as explained in Section \ref{sec:methods}. The scale of the illusions in the Figure is different from the scale used in the experimenta for displaying purposes. The magenta and cyan lines represent the location in the images of the profiles plotted in the rows 2-4.}
\label{fig:grayVI}
\end{figure*}

\begin{figure*}
\begin{center}
\includegraphics[width=\linewidth]{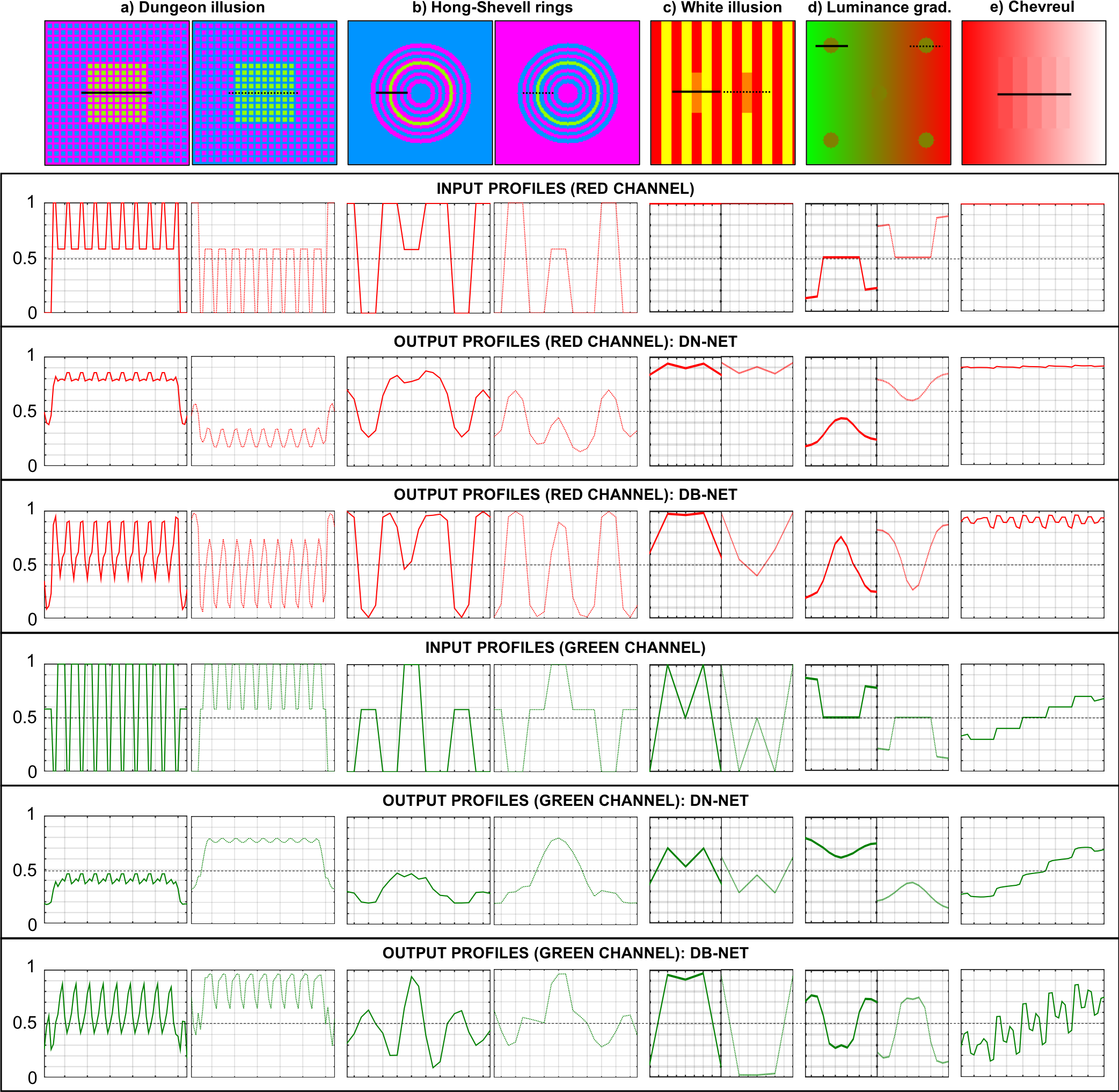}
\end{center}
   \caption{The first row displays the selected color visual illusions as explained in Section \ref{sec:methods}. The scale of the illusions in the Figure is different from the scale used in the experimenta for displaying purposes. The black continuous and dashed lines represent the location in the images of the profiles plotted in the rows 2-4. Only the profiles from the Red and the Green Channels are displayed.}
\label{fig:colorVI}
\end{figure*}

The first row in Figure \ref{fig:grayVI} shows the visual illusions we have selected, with color versions in the top row of \ref{fig:colorVI}.
They are all classical examples of brighntess and color illusions. The illusions  \ref{fig:grayVI}(a) to   \ref{fig:grayVI}(d) presents targets that have identical values but that are seen different depending on their surrounds: in the Dungeon illusion ~\cite{bressan2001explaining} the targets are the large central squares, in Hong-Shevell ~\cite{hong2004brightness} they are the middle rings, in the White illusion ~\cite{white1979new} the targets are the small grey bars, and in the Luminance gradient illusion (combination of~\cite{bruke,adelson2000zj} ) the targets are the circles. The fact that the targets have indeed the same values (0.5 in all cases) can be seen in the second row of Fig.  \ref{fig:grayVI}, that plots the image values along some segments shown in color over the visual illusions in the top row. The Chevreul illusion  ~\cite{ratliff1965mach} presents homogeneous bands of increasing intensity, from left to right, but these bands are perceived to be in-homogeneous, with darker and brighter lines at the borders between adjacent bands.
For color, Fig. \ref{fig:colorVI}, the phenomena are similar: for the Dungeon and the Hong-Shevell cases, the right target must go towards green and the left target towards red; for the White illusion, the left target must go towards yellow and the right target towards red; in the Luminance gradient illusion, the left targets should move towards red and the right targets towards green; finally, the Chevreul illusion should be perceived as in the grayscale case, altough now in the red channel). 


As for the CNNs, we have chosen a very simple architecture and three basic tasks: denoising, deblurring, and color constancy. The architecture has input and output layers of size $128\times128\times3$ pixels,  one hidden layer with eight features maps with a receptive field (kernel size) of five and no stride, and non-linearities (sigmoid activation functions). 
At the end there is a convolutional layer which works as output layer (hence it has three layers for the red, green, and blue channels). Note that no pooling, residual connections or other modifications were added to this architecture. Mean squared error was used as loss function in all the tasks and all the models were implemented\footnote{source code will be made publicly available} using Keras~\cite{chollet2015keras}.  

The tasks are key image processing problems that have close correlates in human perception: denoising relates to our ability to discount noise in images \cite{MCILHAGGA20042659}, deblurring to our capabilities of  avoid perceiving the blur provoked by moving objects \cite{humansdeblurring}, and color constancy to the way our perception of colors matches quite well the reflectance properties of objects and is rather independent of the color of the illuminant.

For denoising we use the Large Scale Visual Recognition Challenge 2014 CLS-LOC validation dataset~\cite{russakovsky2015imagenet} (which contains 50k images), and corrupts images with additive Gaussian noise of $\sigma = 25$ after resize them to 128x128.
For deblurring we use the same dataset as before, and blur the images with a Gaussian kernel of $\sigma = 2$.
For color constancy we have used the dataset of Cheng \emph{et al.} \cite{Cheng14}  that provides the color of the illumination for each image. 
We computed the ground-truth image by applying the inverse of the illuminant color to the original image, and then we performed an end-to-end training between the original image and the ground-truth one. 
For this problem, we divide each original image into four sub-images in order to increase the pool of available images for the training of the net. By doing this, we end up with a total of 6944 images.
In all three cases, the dataset was split in 70\% for training, 20\% for validation, and 10\% for test.

The CNNs are named based on the task they were trained for. Hence, DN-NET, CC-NET, and DB-NET correspond to denoising, color constancy, and deblurring,  respectively.

\section{Experiments and results}

The central experiment of this paper consists in evaluating the performance of DN-NET, CC-NET and DB-NET in the selected grayscale and color visual illusions (see subsection \ref{sec:main_exp}). Then, the influence of the scale of the visual illusion (VI for short) and the size of the receptive field of the CNNs is studied in \ref{sec:oneLayer_scale}. The following subsection (\ref{sec:arch_changes}) shows the effect produced by different variations of the architecture on VI replication. Finally, newer and more complex architectures are considered (see \ref{sec:soa}).
Due to the vast number of combinations of the scale of the VI, the size of receptive fields, and the variations on the architecture of the CNN studied here, the experiments in subsections \ref{sec:oneLayer_scale}, \ref{sec:arch_changes} and \ref{sec:soa} are limited to the denoising task (i.e. variations of DN-NET) and two visual illusions that represent the well-known assimilation (Dungeon) and contrast (Luminance gradient) effects. The figures in the main document only show the most relevant findings from the different variations studied. Please, see the Appendix C for extended versions of the figures presenting the rest of results highlighted in this paper.

\subsection{CNNs replication using low level tasks}\label{sec:main_exp}

In all the results reported in this subsection, both for grayscale and color, the following target sizes were used: Dungeon ($4\times 4$ pixels), Hong-Shevell (ring width of $1$ px.), White ($4\times 4$ px.), Luminance gradient ($5\times 5$ px.) and Chevreul (step width of 10 px.). These are the baseline VI scales used in this paper, hence when referring later to larger or smaller scales is always relative to these baselines.

\subsubsection{Grayscale}\label{sec:gray}

Fig.~\ref{fig:grayVI} shows the results of passing the grayscale VIs through the CNNs trained to perform denoising, color constancy and deblurring. The output profiles showed in Fig.~\ref{fig:grayVI} are the grayscale values obtained using the formula $0.2989·\text{R} + 0.5870·\text{G} + 0.1140·\text{B}$, with R, G and B being the corresponding values in the red, green and blue channels. We can see that DN-NET is capable of replicating illusions from (a) to (d) (see the row \textit{Output profiles: DN-NET} in Fig.~\ref{fig:grayVI}). While Dungeon (a) and H.S. (b) are very well replicated, in White (c) and Lum. (d) the effect is less marked. CC-NET replicates illusions from (b) to (d) (see \textit{Output profiles: CC-NET} in Fig.~\ref{fig:grayVI}) but produces the opposite effect to that of human perception in (a). Finally, DB-NET replicates illusions from (b) to (e), but it presents the same opposite effect as CC-NET in (a). Nevertheless, DB-NET is the only one able to replicate the effect for the Chevreul illusion in grayscale (e).

\subsubsection{Color}\label{sec:color}

In Fig.~\ref{fig:colorVI} we present the results of passing the color VIs through DN-NET and DB-NET. Due to space limitations we omit here the results obtained for CC-NET, which can be found in the Appendix C. Furthermore, the colors for the VIs were chosen in such a way that the most significant results could be observed in the red and green channels, which are the ones chosen to be displayed in Fig.~\ref{fig:colorVI}. We refer again to the Appendix C, where the complete results for DN-NET, DB-NET and CC-NET are displayed in separate figures for the blue (Fig.~\ref{fig:4extB}), the green (Fig.~\ref{fig:4extG}) and the red (Fig.~\ref{fig:4extR}) channels.

DN-NET replicates illusions (a),(b),(c), and (e). For Dungeon (a) and H.S. (b) the right target increases its green value (w.r.t. the input) while the left target increases its value in the red channel. For White (c), the left target gets closer to a yellow color by increasing its green channel value. In the case of Chevreul (e), there is a slight replication in the red channel. Finally, in the case of Lum. (d), DN-NET fails to reproduce the VI.

DB-NET replicates all illusions except for H.S. (b). For illusions Dungeon (a) and White (c) the effect is the same as that observed for DN-NET. For Chevreul (e), the effect is replicated both in the red and the green channels (and also in the blue channel, see Fig.~\ref{fig:4extB}).  Finally in Lum. (d), there is a clear increase in the red and the green channels for the left and the right targets respectively, together with a corresponding decrease of the same channels in the opposite target.

\begin{figure*}[ht!]
\begin{center}
\includegraphics[width=0.8\linewidth]{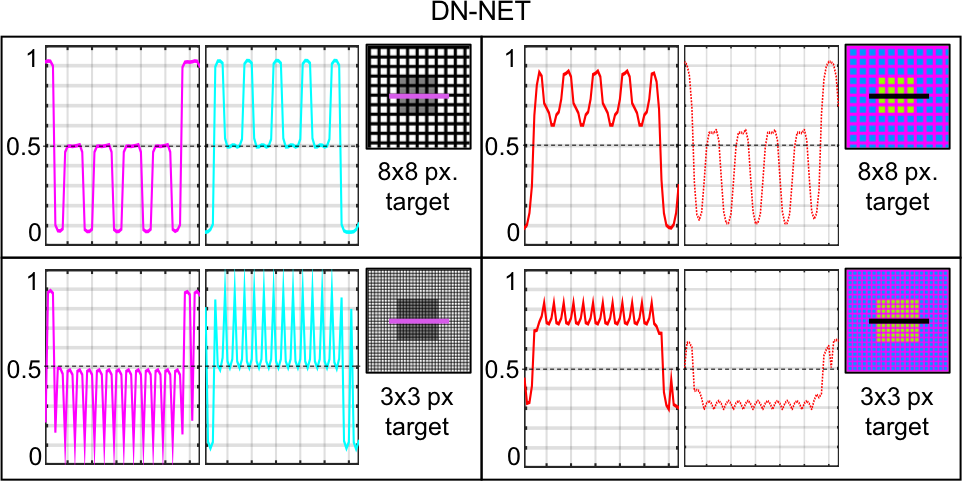}
\end{center}
   \caption{Assimilation results in DN-NET for low and high frequency grayscale and color visual illusions.}
\label{fig:scale}
\end{figure*}
\begin{figure*}[ht!]
\begin{center}
\includegraphics[width=0.8\linewidth]{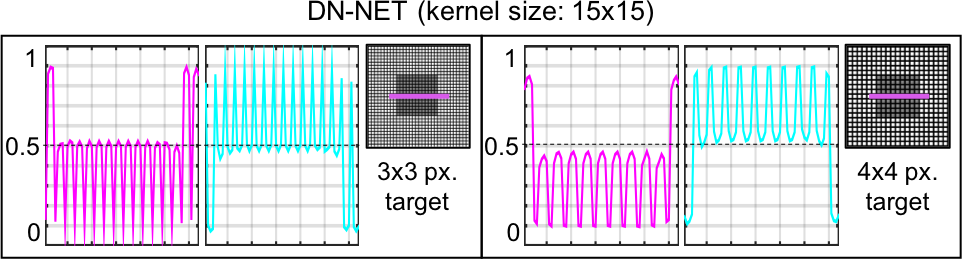}
\end{center}
   \caption{Assimilation results in DN-NET for the largest receptive field (kernel size of $15\times 15$) for the two highest frequencies ($3\times 3$ and $4\times 4$) of the visual illusion.}
\label{fig:kernelsize}
\end{figure*}
\subsection{Influence of the scale of visual illusion and the receptive field size}\label{sec:oneLayer_scale}

In section~\ref{sec:main_exp}, high spatial frequency VIs (i.e. using small patterns) were used to evaluate the response of the CNNs. In humans there is an observed relationship (see e.g. \cite{shapiro2016oxford}) between spatial frequency and visual effect. In most cases this relationship states that higher frequencies imply a larger difference between the observed targets. In this section we study the influence of these changes in the experiments of section (section~\ref{sec:main_exp}). First, the spatial frequency of the VI is reduced in order to study if the same relationship appearing in human perception -where reducing the spatial frequency reduces the replication effect- is observed in the CNNs. Second, we test different receptive field sizes (also called kernel sizes) in the whole architecture, moving from the one used in DN-NET, DB-NET and CC-NET ($5\times 5$) to $3\times 3$, $7\times 7$, $11\times 11$ and $15\times 15$.

\subsubsection{Changing spatial frequency of VI}\label{sec:scales}
DN-NET reduces the replication error when the size of the pattern is increased (increasing the size of the pattern is equivalent to reducing the spatial frequency), therefore emulating the behaviour observed in human perception ~\cite{shapiro2016oxford}). However, the reduction of the effect is dependent on the receptive field size and on whether the illusion is in grayscale or color.

The replication effect observed for DN-NET in the Dungeon illusion in grayscale is completely lost when moving to sizes equal or larger than $8$ pixels (see the left column of the first row in Fig.~\ref{fig:scale}). However, the same VI in color still replicates the effect for that size specially in the red channel (right column of the first row in Fig.~\ref{fig:scale}). The same evolution but in a smaller degree is also observed in the case of Lum. (first row of Fig.~\ref{fig:5ext}).

Furthermore, increasing the spatial frequency leads to an attenuated replication, contrary to the effect produced in human perception. Figure~\ref{fig:scale} in its second row shows how the assimilation effect in Dungeon almost disappears in grayscale. That is also the case for the contrast effect in Lum. (second row of Fig.~\ref{fig:5ext}). In the case of color, the assimilation effect is still clearly present (Fig.~\ref{fig:scale}) but not the contrast effect of Lum. (Fig.~\ref{fig:scale}).


    \subsubsection{Increasing the receptive field}\label{sec:kernels}

A reasonable assumption would be the existence of a relation between the receptive field and the spatial frequency of the patterns. The nature of this relation is not directly understood from the current experiments. In most of the combinations of pattern's frequency and size of receptive field tested the qualitative results do not change. 

For the Lum. VI, the use of larger receptive fields lead to an increase of the replication effect (Fig.~\ref{fig:6ext}). However, for the Dungeon effect when using the largest receptive field size ($15\times 15$), moving from a target size of 4 to 3 pixels changes the assimilation  into a contrast effect (see Fig.~\ref{fig:kernelsize}). For the color VIs there were no significant qualitative changes for either illusion.

\subsection{Variations in architecture and their effect in the replication}\label{sec:arch_changes}

\begin{figure*}[htp!]
\begin{center}
\includegraphics[width=0.8\linewidth]{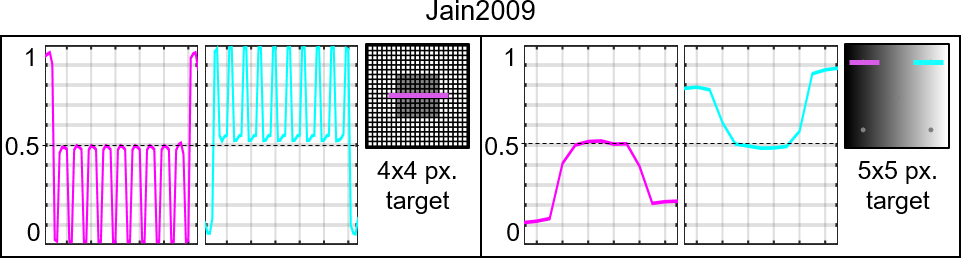}
\end{center}
   \caption{Replication results for Dungeon (assimilation) and Lum. (contrast) for Jain2009, the arquitecture based in  \cite{jain2009natural}.}
\label{fig:jain2009}
\end{figure*}

The purpose of this subsection is to study how common variations in the architecture of CNNs affect the replication of VIs. In order to do so we move from DN-NET to a similar CNN presented by Jain et al. \cite{jain2009natural}, one of the first successful CNNs designed for the purposes of image denoising. Our implementation of this CNN, that we denote as Jain2009 from now on, has an input/ouput size of $128\times 128$ and is composed of four hidden layers with a kernel size of five and a sigmoid as activation function. This CNN can be considered as a deeper version of our DN-NET. We use Jain2009 as the base CNN to be modified with pooling layers, dilated convolutions, and residual connections independently. The training of this CNN followed the same procedure as that of DN-NET.
 Figure ~\ref{fig:jain2009} shows the results of Jain2009. We find replication (although reduced) of both effects in grayscale. Despite being four times deeper than DN-NET, Jain2009 shows qualitatively similar results to the original DN-NET. For the case of color VIs (displayed in Fig.~\ref{fig:7ext}), there is still a replication of the assimilation effect in Dungeon but not of the contrast effect in Lum. (as was the case for DN-NET, see Fig.~\ref{fig:colorVI}). 

\begin{figure*}[htp!]
\begin{center}
\includegraphics[width=0.8\linewidth]{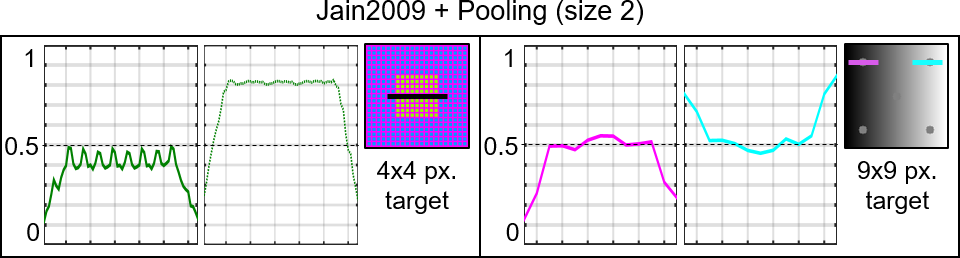}
\end{center}
   \caption{Selected esults from Jain2009 when adding Pooling of size two. Assimilation effect is only replicated in color while the contrast effect is replicated in grayscale.}
\label{fig:jain2009_pool}
\end{figure*}
\subsubsection{Adding pooling layers}\label{sec:jain_pool}

Two pooling layers were added to the Jain2009 architecture: one pooling layer after the first convolutional layer and another after the second convolutional layer. In order to recover the original scale of the input, after the last two hidden convolutional layers an upsampling layer was added.

When pooling layers were used (in this case of  size two) the side effect in grayscale images is that higher frequency VIs are destroyed. Also, in the case of Dungeon the replication is lost, in fact, the opposite effect is observed (see Fig.~\ref{fig:8ext}). However, there is still a replication effect in Lum. for bigger target sizes (see the right column in  Fig.~\ref{fig:jain2009_pool}). In the case of color, the same effect of spatial pattern destruction occurs, but the replication effect still remains for the Dungeon VI in the red and green channels (see the green channel in Fig.~\ref{fig:jain2009_pool}, left column and the red in (Fig.~\ref{fig:8ext}). Larger pooling sizes lead to a total spatial destruction of the patterns in the VIs such that further analysis is prevented.

\subsubsection{Adding Dilated Convolutions}\label{sec:dilated}

\begin{figure*}[htp!]
\begin{center}
\includegraphics[width=0.8\linewidth]{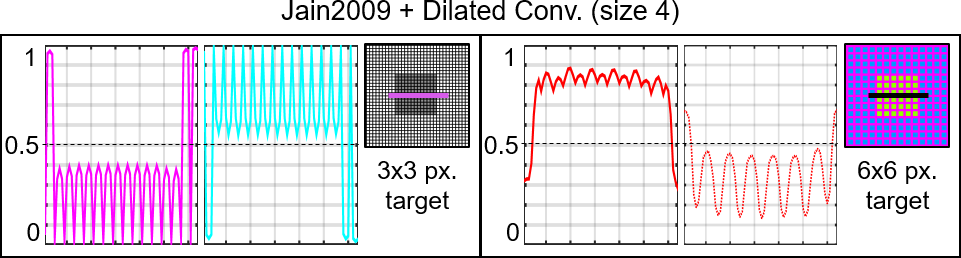}
\end{center}
   \caption{Selected results from Jain2009 when adding Dilated Convolution of size 4. In grayscale the assimilation effect is only present in the highest frequency while in color is visible for lower frequencies.}
\label{fig:jain2009_dilated}
\end{figure*}

For this test, the standard convolutional layers of Jain2009 were replaced with convolutional layers with a dilation rate of 2, 4, and 8.
Two main effects are observed when dilated convolutions were added. First, the contrast effect of Lum. is not replicated in grayscale or color for any of the dilation rates. Second, in all the cases the effect in grayscale for Dungeon, when considering targets equal or larger than 4 pixels, is no longer replicated (Fig.~\ref{fig:9ext}). In fact, it shows a contrast effect instead. However, in the case of color there is still replication for the Dungeon VI even when larger targets are considered (see the right column in Fig.~\ref{fig:jain2009_dilated}).

A special case is observed only when using a dilation of size four: Replication does appear in Dungeon in grayscale for the smallest pattern size (shown in the left column of Fig.~\ref{fig:jain2009_dilated}). This is not the case for any other size of the dilation. 

\subsubsection{Adding Residual Connections}\label{sec:residual}

\begin{figure*}[htp!]
\begin{center}
\includegraphics[width=0.8\linewidth]{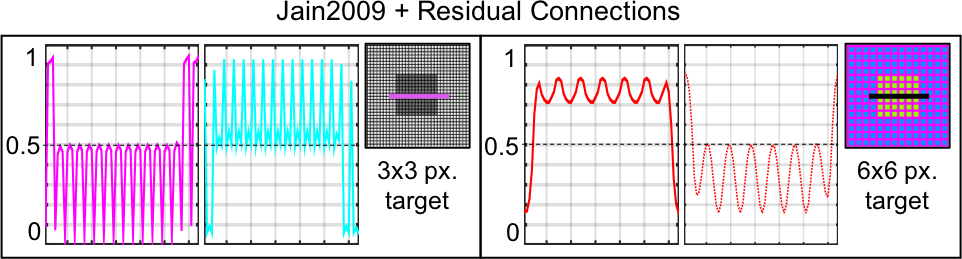}
\end{center}
   \caption{Selected results from Jain2009 when adding residual connections. Assimilation in grayscale is only reproduced in the smallest scale while for color larger scales still produce an effect.}
\label{fig:jain2009_residual}
\end{figure*}

Several configurations of Jain2009 with residual connections were tested. They shared the effect of annulling the replication of both Dungeon and Lum. in grayscale. However, an architecture with a single residual connection going from the output of the first convolutional layer to the input of the final output layer (see Fig.~\ref{fig:CNNResidualJean09}) was still able to  replicate the assimilation effect in the grayscale Dungeon VI for the highest frequency (see the left column in  Fig.~\ref{fig:jain2009_residual}). In the case of color, for all the different variations of Jain2009 with residual connections, there is a replication for Dungeon even if we increase the pattern size (right column in Fig.~\ref{fig:jain2009_residual}) but not for Lum. in none of the cases.

\subsection{Replication in the-state-of-the-art CNNs for image denoising}\label{sec:soa}

\begin{figure*}[htp!]
\begin{center}
\includegraphics[width=0.8\linewidth]{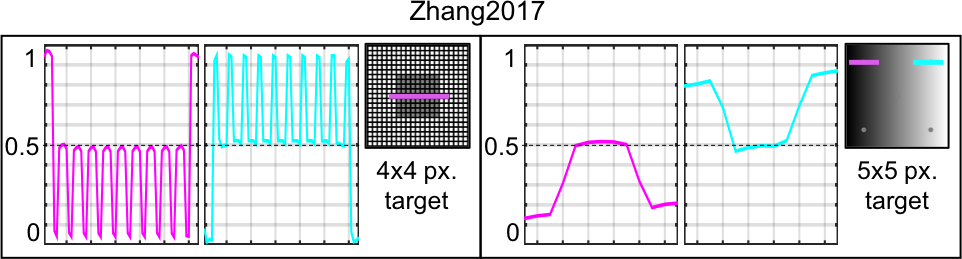}
\end{center}
   \caption{Replication results for Dungeon (assimilation) and Lum. (contrast) for Zhang2017, the state-of-the-art CNN for image denoising presented in \cite{zhang2017beyond}.}
\label{fig:zhang2017}
\end{figure*}

The CNN architectures considered in all previous sections are simple in comparison with modern architectures. Hence, for this test, we choose the best CNN (to the best of our knowledge) for image denoising~\cite{zhang2017beyond} as a model to try to reproduce visual illusions (which we denote as Zhang2017 from now on). Figure ~\ref{fig:zhang2017} shows how, to a small degree,  Zhang2017 can replicate the effect in both Dungeon and Lum. VIs. That is also the case for the color VI (see Fig. ~\ref{fig:11ext}).

\section{Conclusions}

In this work we showed that CNNs trained on natural image databases for basic low-level vision tasks reproduce the human response to visual illusion images, i.e. the CNNs are deceived by the visual illusions in the same way that we are deceived by them. Versions of a  single hidden layer CNN trained for denoising, color constancy, and deblurring were tested to replicate five common visual illusions. Deeper architectures and their common modifications (such as pooling layers, dilated convolutions, and residual connections) were explored too in order to evaluate their effect in the replication of visual illusions. It was found that even the simplest single hidden layer with 8 kernels is already capable of replicating the human response to several grayscale and color illusions. Furthermore, changes in the input image or CNN architecture lead to a change in the illusions that the network is able to reproduce.

We argue that the CNNs in this paper reproduce visual illusions as a by-product of the low level vision tasks of denoising, color constancy or deblurring. Albeit clearly different, the biological correlates of all of these tasks aim to improve the efficiency of the representation and the visual processing, so this supports the argument that visual illusions are the price we have to pay in order to optimally use the limited resources of our visual system.

The illusions that the  CNNs are able to replicate depend on the task each CNN is solving. It would be interesting, from a vision science perspective, to use this insight to try to associate specific illusions (or families of illusions) with visual processing tasks.
Another interesting finding was that CNNs trained with color images can replicate visual illusions in grayscale too: this could maybe give some cues towards answering the question of where precisely in the visual system is the brightness percept derived from color signals, which is still an open one.

Finally, and from a computer vision perspective, if we want CNNs that better replicate human behaviour, we should maybe start aiming for them to better replicate visual illusions.
We are currently working along these lines, developing a CNN architecture with the goal of reproducing as many visual illusions as possible, with validations from psychophysical data.

As future work we want to evaluate if CNNs that replicate visual illusions are more resistant to adversarial attacks that do not fool humans. And to generate new visual illusions using for instance generative adversarial networks.

\section*{Acknowledgements}

This work has received funding from the EU Horizon 2020 programme under grant agreement 761544 (project HDR4EU) and under grant agreement 780470 (project SAUCE) and by the Spanish government and FEDER Fund, grant ref. TIN2015-71537-P (MINECO/FEDER,UE). We gratefully acknowledge the support of NVIDIA Corporation with the donation of the Titan Xp GPU used for this research.

\bibliographystyle{ieee}
\bibliography{arXiV_CNN_2019}

\clearpage

\counterwithin{figure}{section}
\begin{appendices}
\section{CNN architectures}

In this section are depicted the CNN architectures used in the paper. Note that ``Jain2009 + Dilated convolutions'' is not showed here because it is equivalent to Jain2009 with modified convolutional layers.

\subsection{DN-NET, CC-NET, DB-NET}

\begin{figure}[htp!]
\begin{center}
\includegraphics[width=0.9\linewidth]{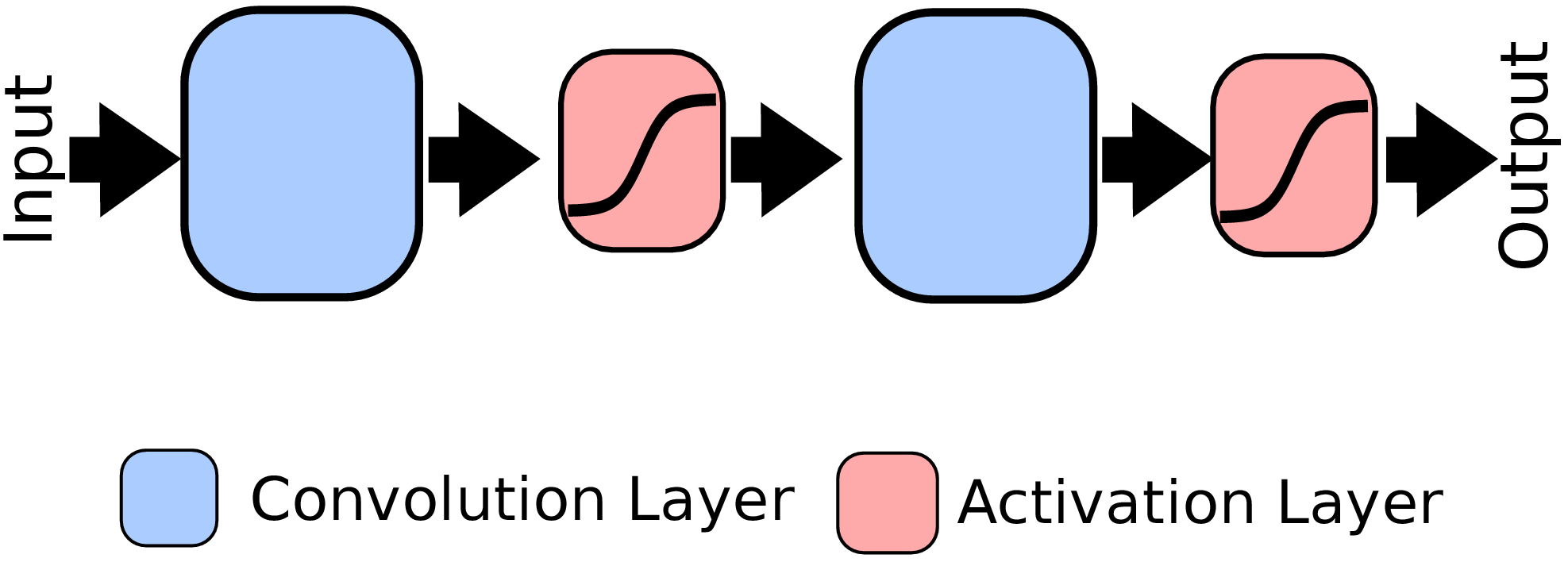}
\end{center}
   \caption{Architecture used in DN-NET, CC-NET, DB-NET}
\label{fig:CNN1layer}
\end{figure}

\subsection{Jain2009}

\begin{figure}[htp!]
\begin{center}
\includegraphics[width=0.9\linewidth]{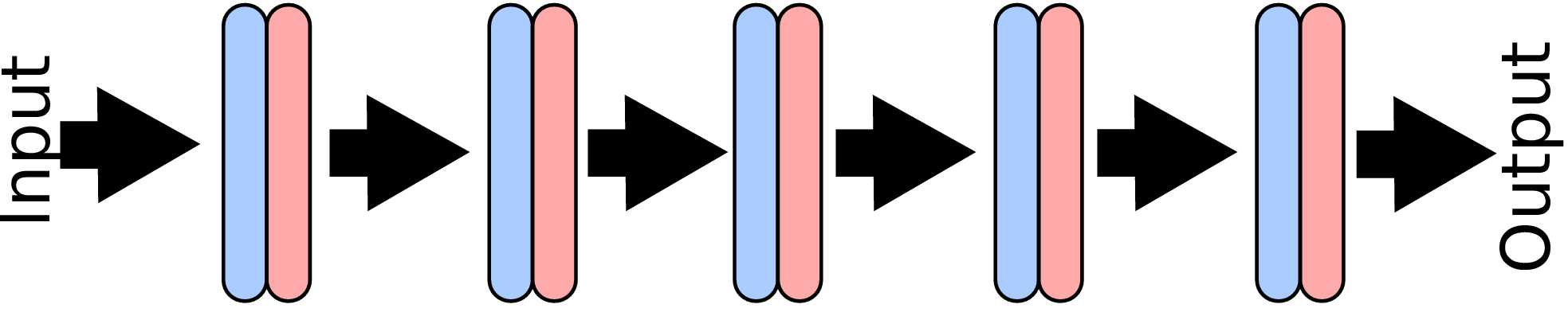}
\end{center}
   \caption{Architecture of Jain2009}
\label{fig:CNNJean09}
\end{figure}

\subsection{Jain2009 + Pooling}

\begin{figure}[ht]
\begin{center}
\includegraphics[width=0.9\linewidth]{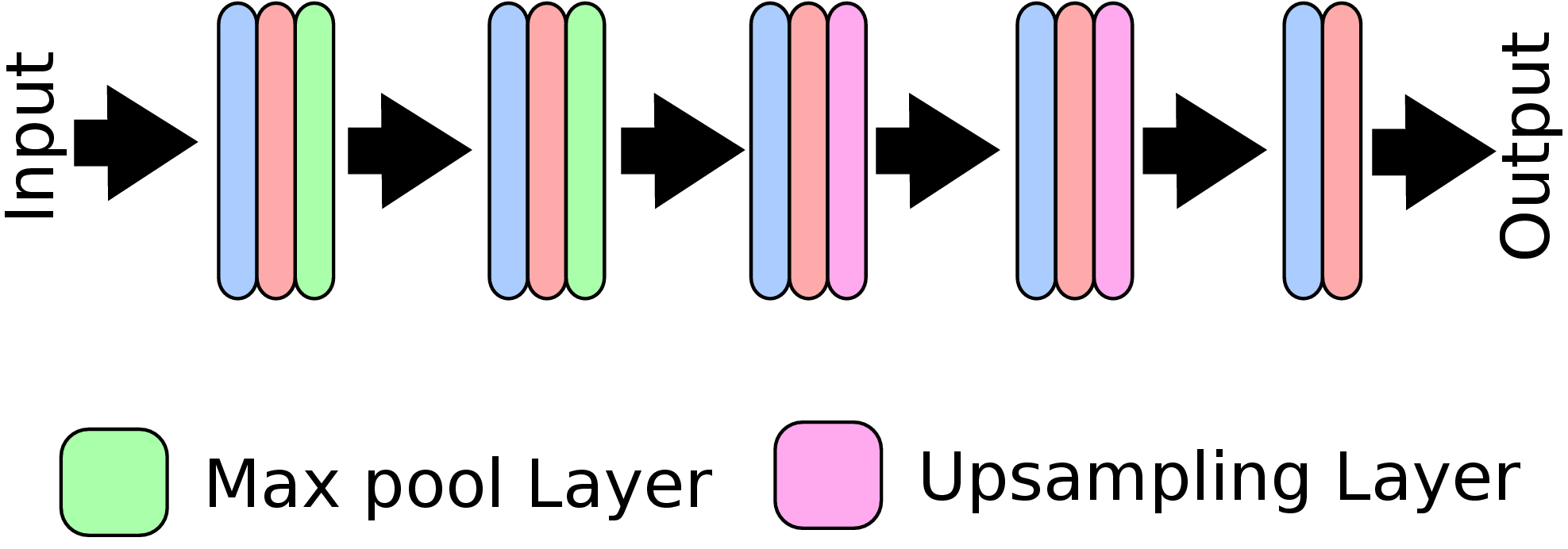}
\end{center}
   \caption{Architecture of Jain2009 + Max pooling }
\label{fig:CNNPoolingJean09}
\end{figure}

\subsection{Jain2009 + Residual connections}

\begin{figure}[htp!]
\begin{center}
\includegraphics[width=0.9\linewidth]{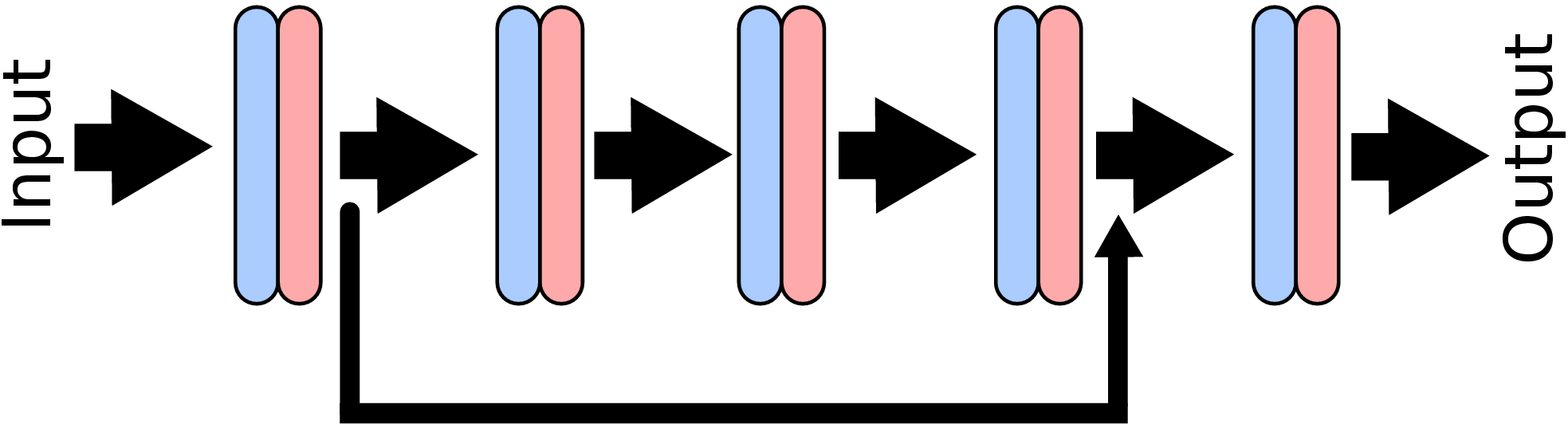}
\end{center}
   \caption{Architecture of Jain2009 + Residual connections}
\label{fig:CNNResidualJean09}
\end{figure}

\section{Implementation details}

All our CNN were trained in Keras framework\cite{chollet2015keras} using mean squared error as loss function and Adam optimizer. The maximum number of epochs was set to 100 and with a batch size of 32. The training stops if there is no improvement in the validation set after two consecutive evaluations.

\section{Extended version of the Figures}

Table~1 shows the correspondence between the Figures presented in the paper (\textbf{Paper Fig.}) and its extended version included in this supplementary material (\textbf{Supp. Fig.}) together with the section of the paper (\textbf{Paper Sect}.) in where each figure is discussed.

\begin{table}[ht]
\centering
\begin{tabular}{ccc}
\hline
\textbf{\begin{tabular}[c]{@{}c@{}}Paper Fig.\end{tabular}} & \textbf{\begin{tabular}[c]{@{}c@{}}Supp. Fig.\end{tabular}} & \textbf{\begin{tabular}[c]{@{}c@{}}Paper Sect.\end{tabular}} \\
\hline
Fig.~\ref{fig:colorVI}  & Fig.~\ref{fig:4extB}  & \ref{sec:color}\\
Fig.~\ref{fig:colorVI}  & Fig.~\ref{fig:4extR}  & \ref{sec:color}\\
Fig.~\ref{fig:colorVI}  & Fig.~\ref{fig:4extG}  & \ref{sec:color}\\
Fig.~\ref{fig:scale}    & Fig.~\ref{fig:5ext}   & \ref{sec:scales}\\
Fig.~\ref{fig:kernelsize} & Fig.~\ref{fig:6ext} & \ref{sec:kernels}\\
Fig.~\ref{fig:jain2009} & Fig.~\ref{fig:7ext}   & \ref{sec:oneLayer_scale}\\
Fig.~\ref{fig:jain2009_pool} & Fig.~\ref{fig:8ext} & \ref{sec:jain_pool}\\
Fig.~\ref{fig:jain2009_dilated} & Fig.~\ref{fig:9ext} & \ref{sec:dilated}\\
Fig.~\ref{fig:jain2009_residual} & Fig.~\ref{fig:10ext} & \ref{sec:residual}\\
Fig.~\ref{fig:zhang2017} & Fig.~\ref{fig:11ext} & \ref{sec:soa}\\
\hline
\vspace{0.1cm}
\end{tabular}
\label{tab:corr}
\caption{Correspondence between the Figure presented in the paper and its extended version in this supplementary material}
\end{table}

\begin{figure*}[htp!]
\begin{center}
\includegraphics[width=\linewidth]{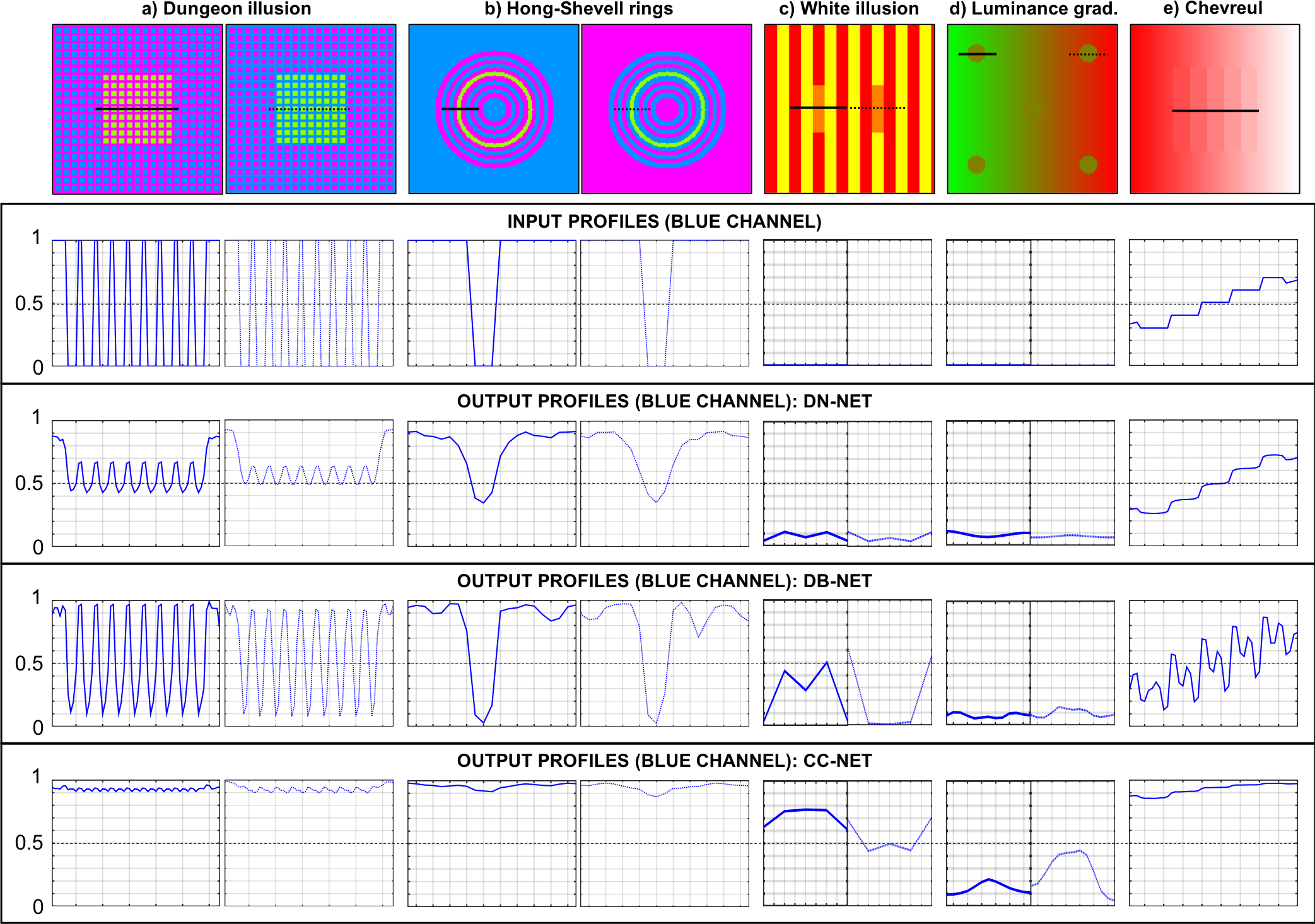}
\end{center}
   \caption{Extension of the Fig.~\ref{fig:colorVI} presenting only the blue channel profiles.}
\label{fig:4extB}
\end{figure*}

\begin{figure*}[ht]
\begin{center}
\includegraphics[width=\linewidth]{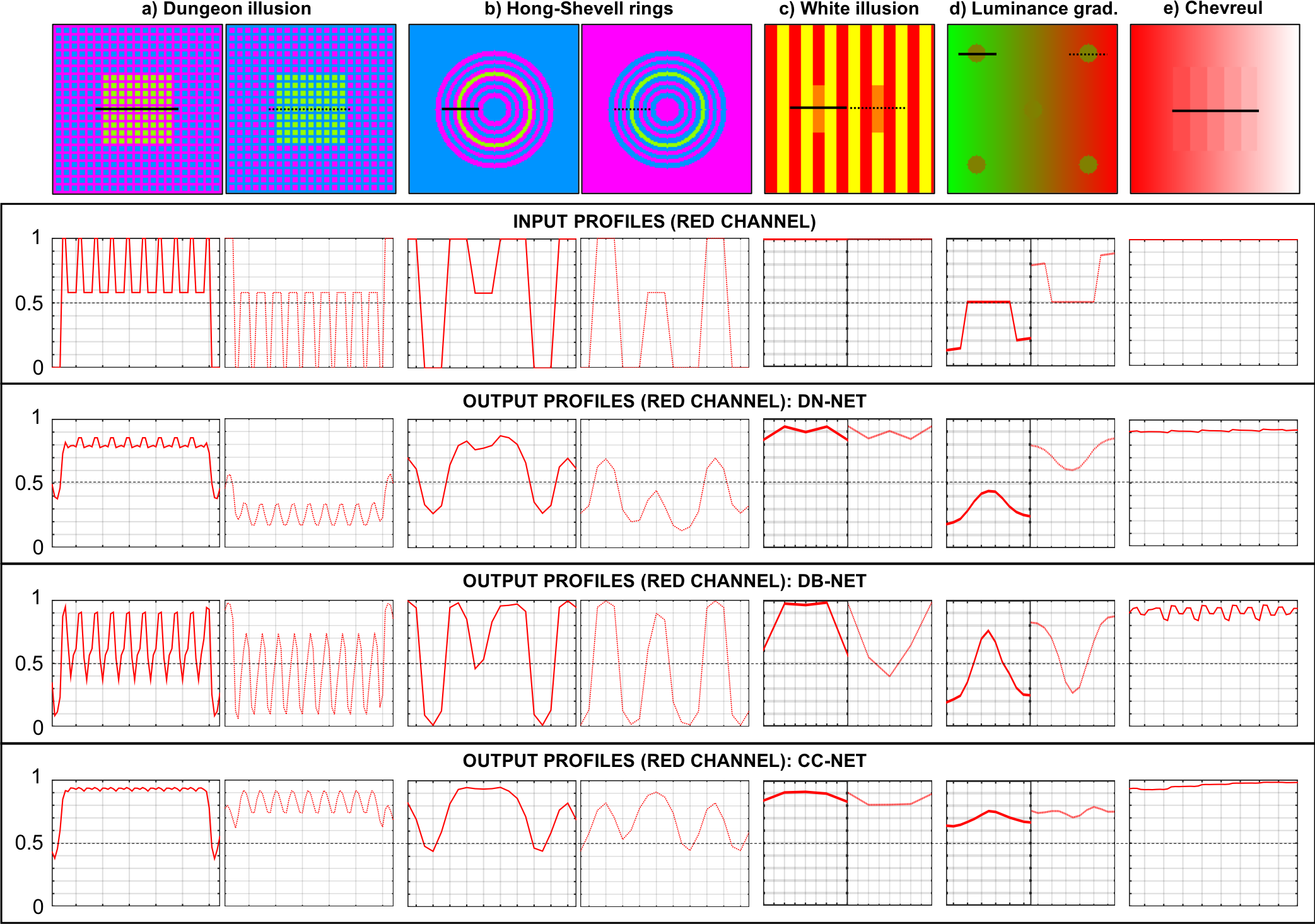}
\end{center}
   \caption{Extension of the Fig.~\ref{fig:colorVI} presenting only the red channel profiles.}
\label{fig:4extR}
\end{figure*}

\begin{figure*}[ht]
\begin{center}
\includegraphics[width=\linewidth]{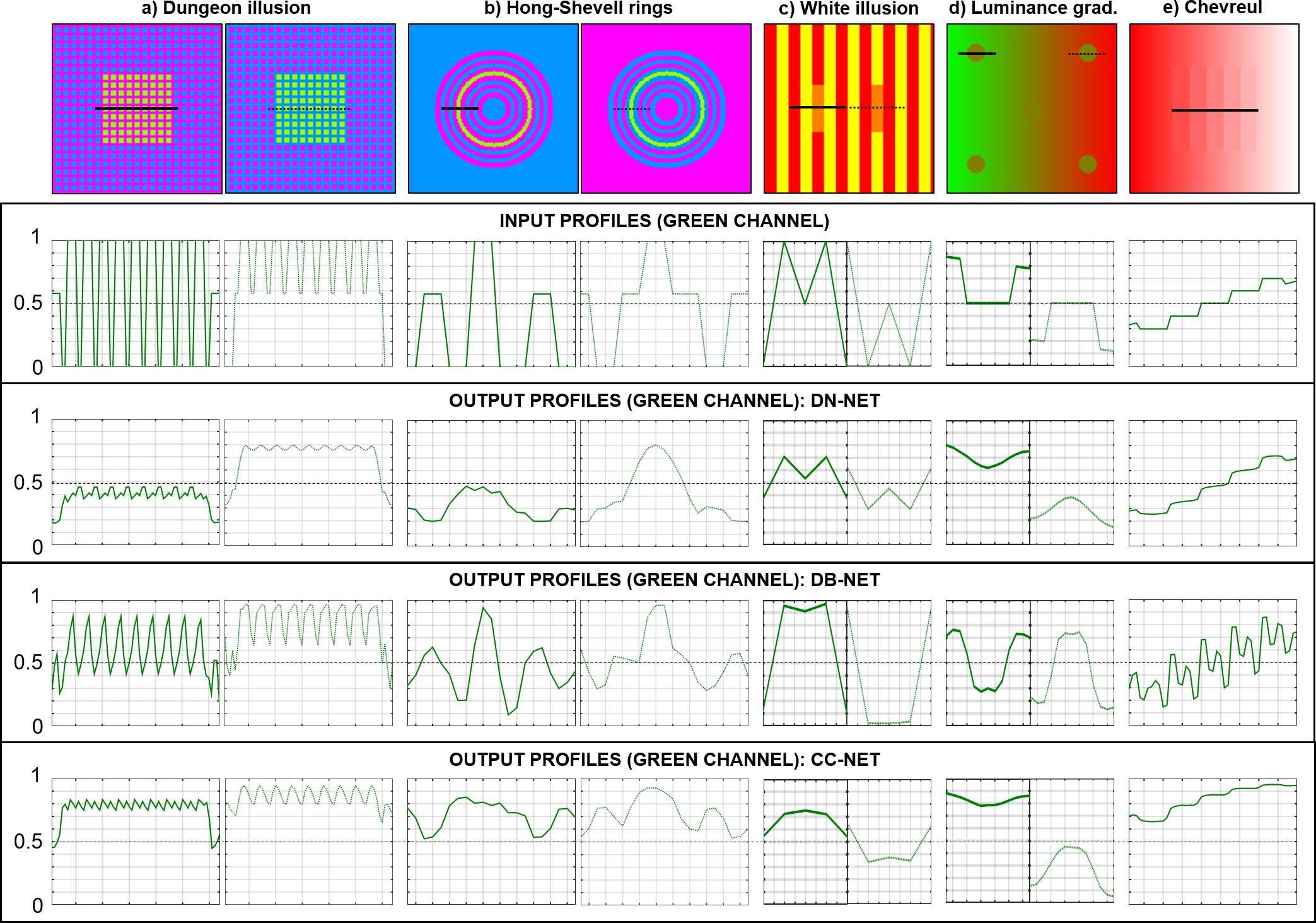}
\end{center}
   \caption{Extension of the Fig.~\ref{fig:colorVI} presenting only the green channel profiles.}
\label{fig:4extG}
\end{figure*}

\begin{figure*}[ht]
\begin{center}
\includegraphics[width=0.8\linewidth]{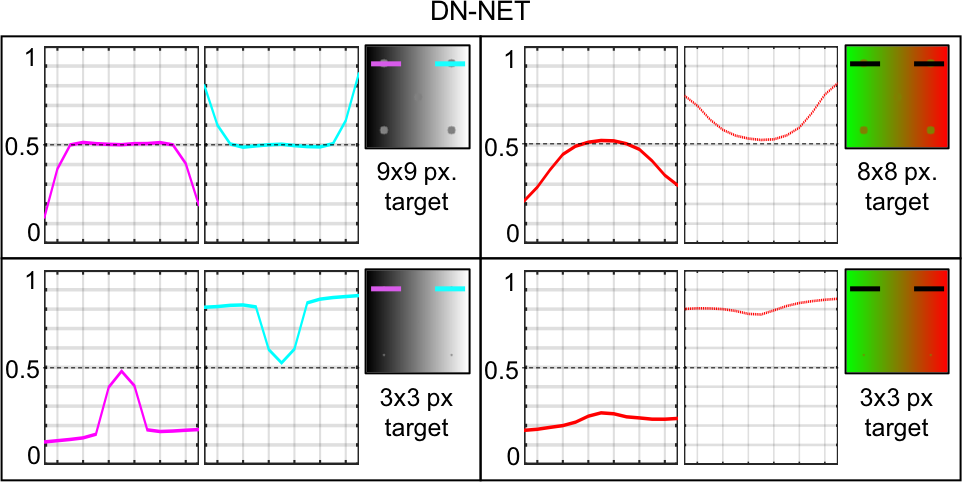}
\end{center}
   \caption{Extension of the Fig.~\ref{fig:scale} showing Lum. in grayscale and color (red channel) for small and large targets.}
\label{fig:5ext}
\end{figure*}

\begin{figure*}[ht]
\begin{center}
\includegraphics[width=0.8\linewidth]{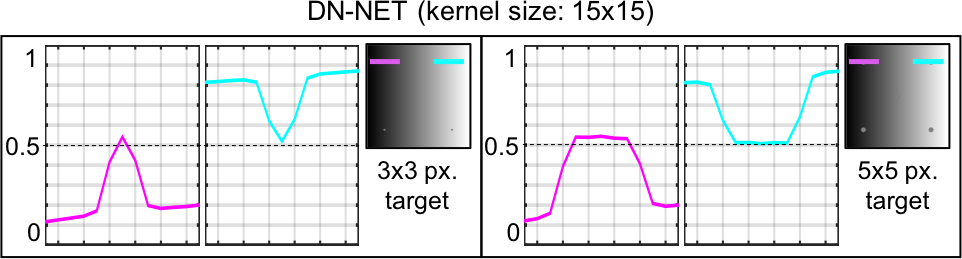}
\end{center}
   \caption{Extension of the Fig.~\ref{fig:kernelsize} presenting Lum. in grayscale for the scale used in Fig.~\ref{fig:colorVI} and a smaller one.}
\label{fig:6ext}
\end{figure*}

\begin{figure*}[ht]
\begin{center}
\includegraphics[width=0.8\linewidth]{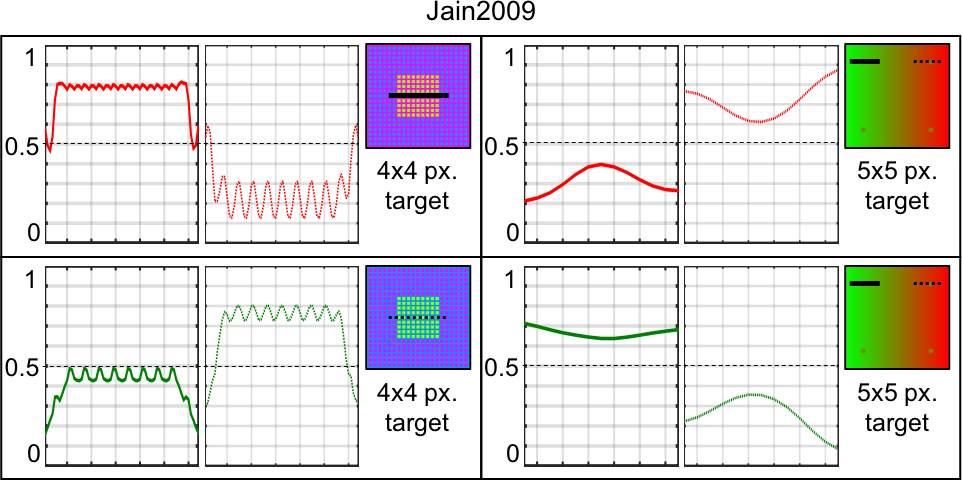}
\end{center}
   \caption{Extension of the Fig.~\ref{fig:jain2009} showing Dungeon and Lum. in color (red and green channels) for the scales used in Fig.~\ref{fig:colorVI}.}
\label{fig:7ext}
\end{figure*}

\begin{figure*}[ht]
\begin{center}
\includegraphics[width=0.8\linewidth]{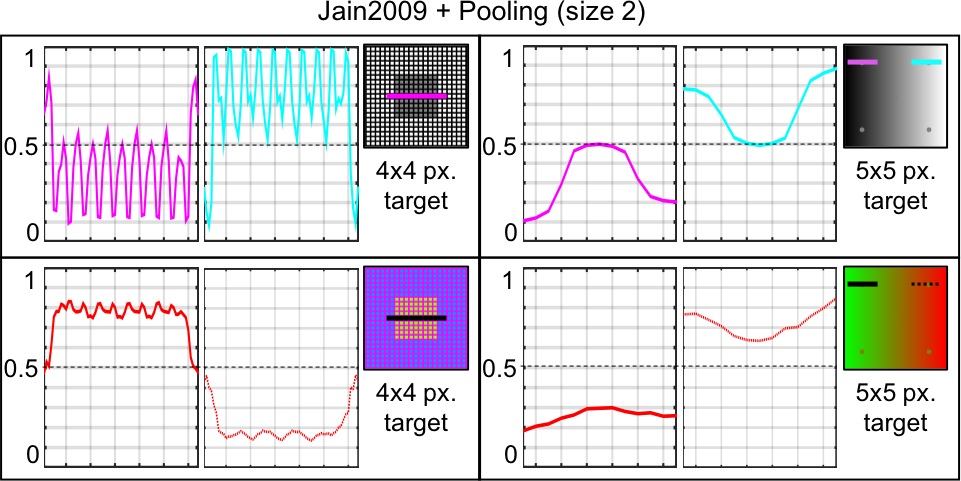}
\end{center}
   \caption{Extension of the Fig.~\ref{fig:jain2009_pool} presenting Dungeon and Lum.  grayscale and color (red channel) in the same scales used in Fig.~\ref{fig:colorVI}.}
\label{fig:8ext}
\end{figure*}

\begin{figure*}[ht]
\begin{center}
\includegraphics[width=0.8\linewidth]{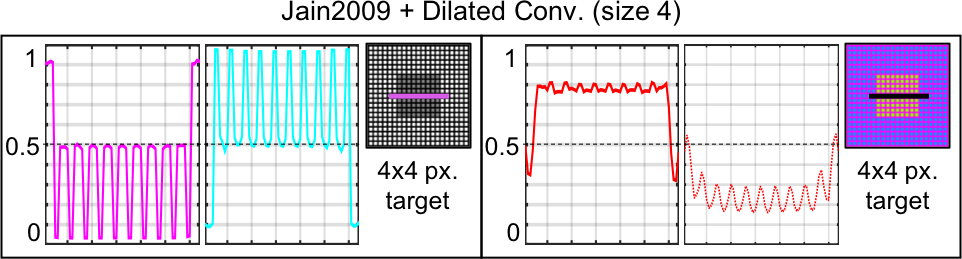}
\end{center}
   \caption{Extension of the Fig.~\ref{fig:jain2009_dilated} presenting Dungeon results in grayscale and color (red channel) using the same scale of Fig.~\ref{fig:colorVI}.}
\label{fig:9ext}
\end{figure*}

\begin{figure*}[ht]
\begin{center}
\includegraphics[width=0.8\linewidth]{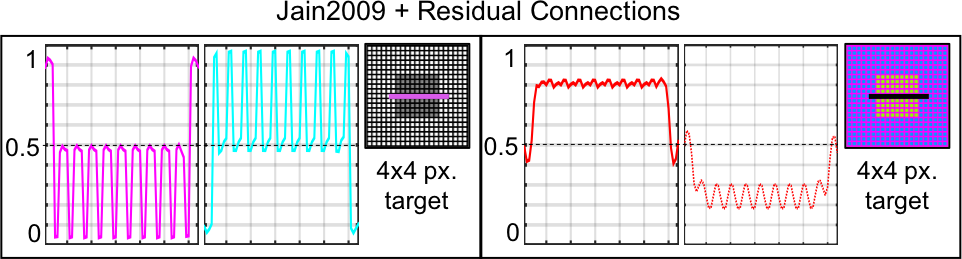}
\end{center}
   \caption{Extension of the Fig.~\ref{fig:jain2009_residual} showing Dungeon results in grayscale and color (red channel) using the same scale of Fig.~\ref{fig:colorVI}.}
\label{fig:10ext}
\end{figure*}

\begin{figure*}[ht]
\begin{center}
\includegraphics[width=0.8\linewidth]{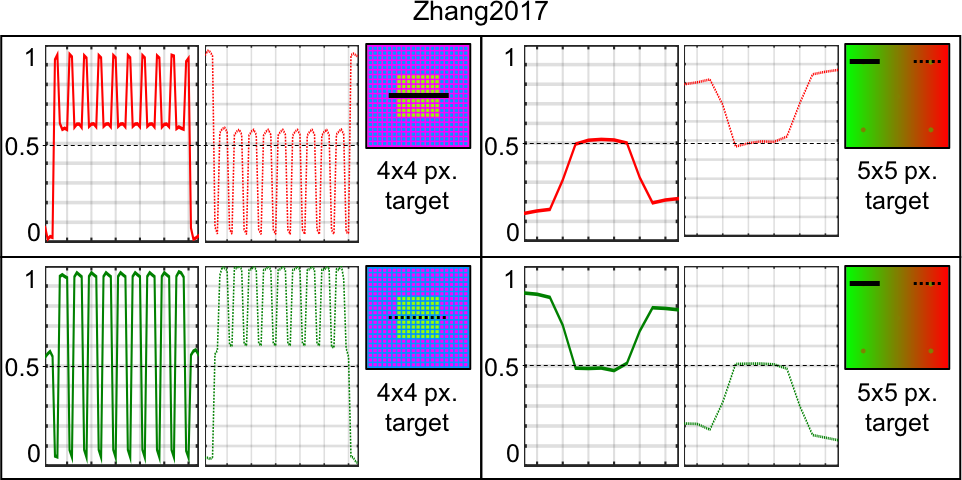}
\end{center}
   \caption{Extension of the Fig.~\ref{fig:zhang2017} presenting Dungeon and Lum. results in color (red and green channels) in the baseline scale of Fig.~\ref{fig:colorVI}.}
\label{fig:11ext}
\end{figure*}

\end{appendices}

\end{document}